\documentclass{article}
\usepackage[accepted]{icml2020}
\usepackage{microtype}
\usepackage{graphicx}
\usepackage{booktabs} 
\usepackage{amsbsy}
\usepackage{amsmath}
\usepackage{amsthm}
\usepackage{mathtools}
\usepackage{bbold}
\usepackage{hyperref}

\usepackage{xcolor}
\usepackage{soul}
\usepackage{cleveref}
\usepackage{xspace}
\usepackage{hhline}
\usepackage{multirow}
\usepackage{dsfont}
\usepackage{amssymb}

\newcommand{\ie}{\textit{i}.\textit{e}., }

\newcommand{\cg}{\mathrm{CG}}

\newcommand{\knn}{\mathrm{kNN}}

\newcommand{\cgnn}{\mathrm{CGNN}}

\newcommand{\troi}{\mathrm{RoI}}

\newcommand{\cgexplainer}{\textsc{CGExplainer}}
\newcommand{\rgexplainer}{\textsc{RGExplainer}}
\newcommand{\explainer}{\textsc{GnnExplainer}}
\newcommand{\model}{\mathcal{M}}
\newcommand{\loss}{\mathcal{L}}
\newcommand{\entropy}{\mathcal{H}}

\DeclareMathOperator{\E}{\mathds{E}}
\DeclareMathOperator{\diag}{\mathrm{diag}}
\newcommand{\kd}{\mathrm{KD}}
\newcommand{\ce}{\mathrm{CE}}
\newcommand{\dist}{\mathrm{DIST}}
\newcommand{\dataset}{\mathcal{D}}

\usepackage{subcaption}
\usepackage{caption}

\begin{document}

\twocolumn[

\icmltitle{Towards Explainable Graph Representations in Digital Pathology}

\icmlsetsymbol{equal}{*}

\begin{icmlauthorlist}
\icmlauthor{Guillaume Jaume*}{ibm,epfl}
\icmlauthor{Pushpak Pati*}{ibm,eth_itet}
\icmlauthor{Antonio Foncubierta-Rodriguez}{ibm}
\icmlauthor{Florinda Feroce}{pascale}
\icmlauthor{Giosue Scognamiglio}{pascale}
\icmlauthor{Anna Maria Anniciello}{pascale}
\icmlauthor{Jean-Philippe Thiran}{epfl}
\icmlauthor{Orcun Goksel}{eth_itet}
\icmlauthor{Maria Gabrani}{ibm}
\end{icmlauthorlist}

\icmlaffiliation{eth_itet}{Department of Information Technology and Electrical Engineering, ETH Z\"urich, Z\"urich, Switzerland}
\icmlaffiliation{epfl}{Institute of Electrical Engineering, EPFL, Lausanne, Switzerland}
\icmlaffiliation{ibm}{IBM Research Z\"urich, Z\"urich, Switzerland}
\icmlaffiliation{pascale}{National Cancer Institute - IRCCS-Fondazione Pascale, Naples, Italy}

\icmlcorrespondingauthor{Guillaume Jaume}{gja@zurich.ibm.com}

\icmlkeywords{Machine Learning, ICML, Digital Pathology GNN, Interpretability}

\vskip 0.3in
]



\printAffiliationsAndNotice{\icmlEqualContribution} 

\begin{abstract}
Explainability of machine learning (ML) techniques in digital pathology (DP) is of great significance to facilitate their wide adoption in clinics. Recently, graph techniques encoding relevant biological entities have been employed to represent and assess DP images. Such paradigm shift from pixel-wise to entity-wise analysis provides more control over concept representation. In this paper, we introduce a post-hoc explainer to derive compact per-instance explanations emphasizing diagnostically important entities in the graph. Although we focus our analyses to cells and cellular interactions in breast cancer subtyping, the proposed explainer is generic enough to be extended to other topological representations in DP. Qualitative and quantitative analyses demonstrate the efficacy of the explainer in generating comprehensive and compact explanations.
\end{abstract}

\section{Introduction}


Convolutional Neural Networks (CNNs), so far the most successful ML approach in image analysis, have been widely adopted to assess DP images to improve diagnosis and patient outcome.
However, concept representations of CNNs remain unexplained in DP and thus hinder their adoption in typical workflows. Therefore, explainable ML technologies in DP have become of paramount interest to build trust and promote the employment of ML in clinical settings~\cite{Holzinger2017}.

Typically CNNs process complex and large DP images in a patch-wise manner, followed by aggregating the patch-wise learning to address downstream DP tasks. 
Recently, several research works have been devoted to demystify the concept representations of CNNs in automated diagnosis. Patch-level explainable methods~\cite{Graziani2018, Hagele2019,Korbar2017, Mobadersany2017, Cruz-Roa2013, Xu2017a} build patch-level \emph{heatmaps}, where an importance score is computed per pixel to identify the regions of importance. For instance,~\citet{Hagele2019} use layer-wise relevance propagation~\cite{Bach2015} to generate positive scores for pixels that are positively correlated with the class label and negative scores otherwise.
Such approaches have several limitations. 
First, pixel-level heatmaps fail to capture the spatial organization and interactions of relevant biological entities. 
Second, the pixel-level analysis is completely detached from any biological reasoning that pathology guidelines recommend for decision making. 
Third, pixel-level explanation are common in the form of blurry heatmaps, which then do not allow to discriminate the relevance of nearby entities and their interactions.

Recently, graph techniques have been adopted to map DP patches to graph representations and process such graphs for pathology tasks~\cite{Gunduz2004,Zhou2019,Sharma2016,Gadiya2019,Wang2019,Pati2020}. Graph representations embed biological entities and their interactions. To the best of our knowledge, explainability of \emph{graph-based} approaches for DP has not been addressed yet.
In this paper, a major step towards explainability in DP is presented based on two proposals: 
First, we advocate for shifting the analysis from a pixel-level representation to a relevant biological entity/relationship-oriented representation. The learning can then be regulated to specific entities and interactions, aligned with the biological and pathological knowledge. 
Second, we propose to adopt an instance-level post-hoc explainability method that extracts a relevant subset of entities and interactions from the input graph.
We define this subset as the explanation of our original graph analysis. 
We hypothesize that the explanation will be deemed useful if and when the subset aligns with prior pathological knowledge.
In this paper, we map DP images to cell-graphs~\cite{Gunduz2004}, where cells and cellular interactions are represented as nodes and edges of the graph, and focus on the interpretability of cell-graphs towards cancer subtyping. 

\section{Methodology}

In this section, we first present the extraction of \emph{graph} representations from DP images, and further present the Graph Neural Network (GNN) framework for the processing of the representations. Second, we introduce the explainability module to acquire comprehensive explanations.

\subsection{Cell-graph representation and learning} 
\label{sec:imgtograph}

\begin{figure}[t]
\centering
\setlength{\belowcaptionskip}{-5pt}
\includegraphics[width=\linewidth]{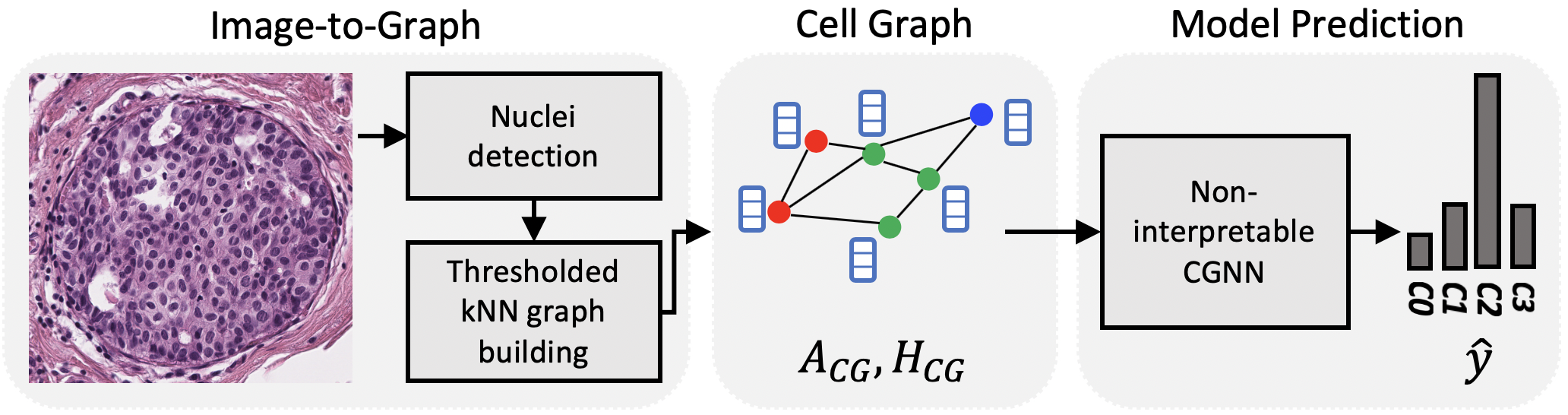}
\captionsetup{width=\linewidth}
\caption{A $\troi$ is transformed into a $\cg$, and is processed by $\cgnn$ to predict the cancer subtype.}
\label{fig:cell_graph}
\end{figure}


The DP images are transformed into cell-graph ($\cg$) representations. Formally, we define a $\cg$, $G_\cg=(V, E, H)$ as an undirected graph composed of a vertices $V$ and edges $E$. Each vertex is described by an embedding $h \in \mathds{R}^d$, or equivalently expressed in its matrix form as $H \in \mathds{R}^{|V| \times d}$. The graph topology is described by a symmetric adjacency matrix $A \in \mathds{R}^{|V| \times |V|}$, where $A_{u,v} = 1$ if an edge exists between vertices $u$ and $v$. 

To build $\cg$, we detect nuclei at 40$\times$ resolution using Hover-Net \cite{Graham2019}, a state-of-the-art nuclei segmentation algorithm pre-trained on MoNuSeg dataset \cite{Kumar2020}. 
We extract 16 hand-crafted features incorporating shape, texture and color attributes to represent each nucleus as in~\cite{Zhou2019}.
We include centroid location normalised by the image size to spatially encode the nucleus.
The detected nuclei and their 18-dimensional embeddings serve as the node and initial node embeddings of our $\cg$.
The $\cg$~topology assumes that spatially close cells encode biological interactions and consequently should form an edge. 
We use the k-Nearest Neighbors ($\knn$) algorithm, \ie for each node $u$, we build edges $e_{uv}$ to the $k$ closest vertices $v$.
As isolated cells have weak cellular interaction with other cells, they ought to stay detached. Thus, we threshold the $\knn$~graph by removing edges that are longer than a specified distance. 
We set $k=5$ and the distance threshold to $50$ pixels in our modeling. 

For the downstream DP task, we determine the breast cancer subtypes of regions-of-interest ($\troi$s). For a dataset with $N~\troi$s, we create $\dataset = \{ G_{\cg, i}, l_i \}_{i=\{1, ..., N\}}$ consisting of $N$ $\cg$s and corresponding cancer stage labels $l_i$. A GNN~\cite{Defferrard2016,Kipf2017, Velickovic2017,Xu2019}, denoted as $\cgnn$, is employed to build fixed-size graph embeddings from the $\cg$s. These embeddings are fed to a Multi-Layer Perceptron (MLP) to predict the cancer stages.
In particular, we use the Graph Isomorphism Network (GIN)~\cite{Xu2019}, an instance of message passing neural network~\cite{Gilmer2017b}. A block diagram with the main steps is presented in Figure~\ref{fig:cell_graph}.

\subsection{Cell-graph explainer}

\begin{figure}[t]
\centering
\setlength{\belowcaptionskip}{-5pt}
\includegraphics[width=\linewidth]{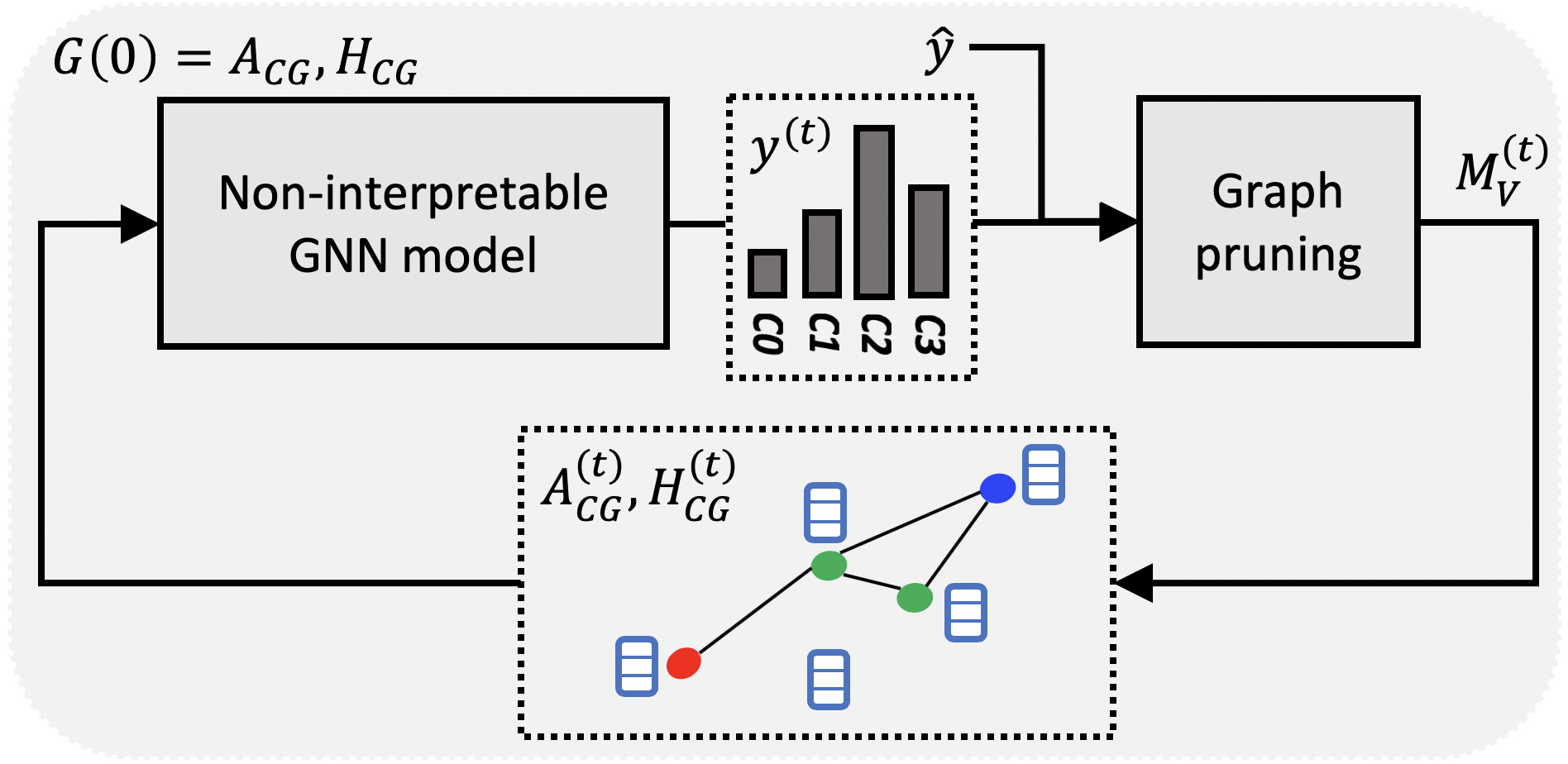}
\captionsetup{width=\linewidth}
\caption{Overview of $\cgexplainer$. The original $\cg$ is iteratively pruned until convergence of the optimization.}
\label{fig:explainer}
\end{figure}

We propose a cell-graph explainer ($\cgexplainer$) inspired by the $\explainer$~\cite{Ying2019}, a post-hoc interpretability method based on a graph pruning optimization. 
Considering the large number of cells in a $\troi$, we hypothetize that many of them will provide little information in the decision making, whereas others will be responsible for class specific patterns that would allow better understanding of the disease.
Thus, we prune the redundant and uninformative graph components, and define the resulting sub-graph as the \emph{explanation}. 

Formally, let us consider a trained GNN model $\model$, 
and a sample $\{G_\cg, l\}$ from $\dataset$ predicted as $\hat{y} = \model(G_\cg)$. We aim to find a sub-graph $G_s=(V_s, E_s, H_s) \subset G_\cg$ such that the mutual information between the original prediction and the sub-graph is maximized, \ie
\begin{align}
    \max_{G_s} \mathrm{MI}(\hat{Y}, G_s) = \entropy(\hat{Y}) - \entropy( \hat{Y} | G_\cg=G_s )
\end{align}
which is equivalent to minimizing the conditional entropy:
\begin{align} \label{eq:explainer2}
    \entropy( \hat{Y} | G_\cg=G_s ) = - \E_{\hat{Y}|G_s}[\log(P_\model(\hat{Y}|G_s))]
\end{align}
Intuitively, $G_s$ maximizes the probability of $\hat{y}$.
Direct optimization of Equation~\eqref{eq:explainer2} is intractable due the combinatorial nature of graphs. Therefore, the $\explainer$ proposes to learn a mask that activates or deactivates parts of the graph. Considering the coherent pathological explainability of cells compared to cellular interactions, we focus on interpreting the \emph{cells} in this work. Thus, we aim at learning a mask $M_V$ at \emph{node-level} that satisfies:
\begin{align} \label{eq:explainer3}
\min_{M_V} - \sum_{c=1}^C \mathds{1}_{[y=c]} \log (P_\model (\hat{Y} | G_\cg,\sigma(\diag(M_V)) H)))
\end{align}
where $C$ denotes the number of classes, $\sigma$ is the sigmoid activation, and $\diag: \mathds{R}^{|V|} \rightarrow \mathds{R}^{|V| \times |V|}$ is the diagonal matrix of the weight vector $M_V$. We intend the explanations to be as compact as possible, ideally with binarized weights, while providing the same prediction as the original graph. 
Heuristically, we enforce these constraints by minimizing:
\begin{align} \label{eq:explainer4}
    \loss = \loss_\kd(\hat{y}, y^{(t)}) + \alpha_{M_V} \sum_{i}^{|V|} \sigma(M_{V_i}^{(t)}) + \alpha_{\entropy} \entropy^{e}(\sigma(M_V^{(t)}))
\end{align}
where, $t$ is the optimization step.
First term is the knowledge-distillation loss $\loss_\kd$ between the new logits $y^{(t)}$ and the original prediction $\hat{y}$. Second term ensures the compactness of $M_V$. Third term binarizes $M_V$ by minimizing its element-wise entropy $\entropy^{e}$.
Following~\cite{Hinton2015a}, $\loss_\kd$ is a combination of distillation and cross-entropy loss:
\begin{align} \label{eq:explainer5}
    \loss_\kd = \lambda \loss_\ce + (1 - \lambda) \loss_\dist \; \text{where} \: \lambda = \frac{\entropy^{e}(y^{(t)})}{\entropy^{e}(\hat{y})}
\end{align}
As the element-wise entropy $\entropy^{e}(y^{(t)})$ increases, $\loss_\ce$ gains importance and avoids a change in predicted label.
$M_V$, produced by optimizing Equation~\eqref{eq:explainer4}, identifies important nodes with a weight factor. An overview of the explainer module is shown in Figure~\ref{fig:explainer}.

\section{Experiments}

\subsection{Dataset}

\begin{table*}
  \centering
   \setlength{\belowcaptionskip}{-5pt}
   \setlength{\tabcolsep}{0.5em}
   \renewcommand{\arraystretch}{1.2}
  \begin{tabular}{l|cc|c|ccc|c|ccccc|c}
    \hline
    \multirow{2}{*}{Metric/Scenario} & \multicolumn{3}{c|}{\textbf{2-class scenario}} & \multicolumn{4}{c|}{\textbf{3-class scenario}} & \multicolumn{6}{c}{\textbf{5-class scenario}}\\
    & {N+B} & {D+I} & {All} & {N+B} & {A} & {D+I} & {All} & {N} & {B} & {A} & {D} & {I} & {All}\\
    \hline
    Weighted F1-score ($\uparrow$)   & $0.97$ & $0.97$ & $0.97$      & $0.95$ & $0.35$ & $0.80$ & $0.77$     & $0.56$ & $0.74$ & $0.37$ & $0.62$ & $0.77$ & $0.61$ \\ 
    \hline
    Node reduction (\%) ($\uparrow$) & $97.7$ & $91.6$ & $94.6$      & $89.5$ & $92.4$ & $85.6$ & $88.5$     & $92.3$ & $93.8$ & $75.8$ & $63.3$ & $59.00$ & $76.9$   \\ 
    Edge reduction (\%) ($\uparrow$) & $99.2$ & $93.8$ & $96.4$      & $94.3$ & $98.7$ & $90.5$ & $93.5$     & $97.1$ & $97.0$ & $90.6$ & $74.1$ & $62.8$ & $84.2$   \\
    \hline
    Original $\ce$ ($\downarrow$)  & $0.21$ & $0.21$ & $0.21$      & $0.45$ & $2.05$ & $0.38$ & $0.72$     & $2.65$ & $0.59$ & $2.22$ & $0.72$ & $0.48$ & $1.21$ \\ 
    Explanation $\ce$ ($\downarrow$) & $0.10$ & $0.21$ & $0.16$      & $0.44$ & $1.41$ & $0.55$ & $0.67$     & $1.65$ & $0.73$ & $1.61$ & $2.57$ & $0.67$ & $1.41$ \\
    Random $\ce$ ($\downarrow$)    & $0.02$ & $3.14$ & $1.61$       & $1.00$ & $0.38$ & $1.75$ & $1.20$     & $0.62$ & $0.93$ & $1.52$ & $11.4$ & $2.85$ & $3.55$ \\ 
    \hline
  \end{tabular}
  \caption{Quantitative results for $\cgnn$, $\cgexplainer$ compactness, $\cgexplainer$ and $\rgexplainer$ performances.}
  \label{table:table1}
\end{table*}

\begin{figure*}[t]
\centering
\setlength{\belowcaptionskip}{-5pt}
\begin{subfigure}{.24\linewidth}
    \centering
    \includegraphics[width=\linewidth]{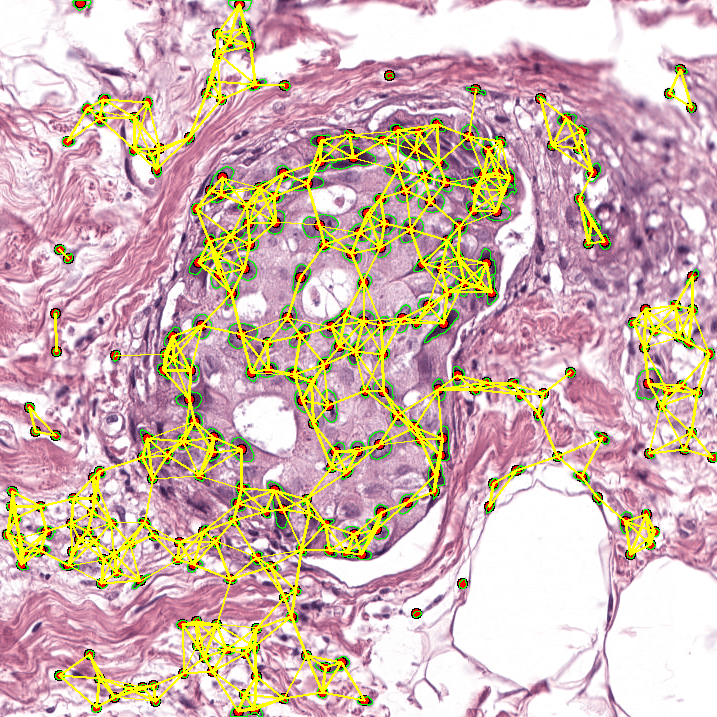}
    \caption{Original $\cg$}\label{fig:image1}
\end{subfigure} 
\hfill
\begin{subfigure}{.24\linewidth}
    \centering
    \includegraphics[width=\linewidth]{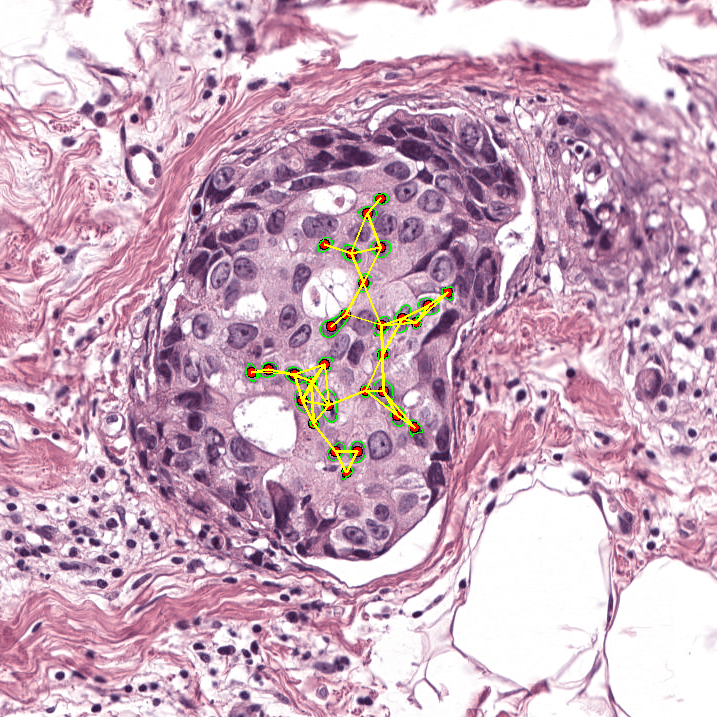}
    \caption{2-class explanation $\cg$}\label{fig:image2}
\end{subfigure}
\hfill
\begin{subfigure}{.24\linewidth}
    \centering
    \includegraphics[width=\linewidth]{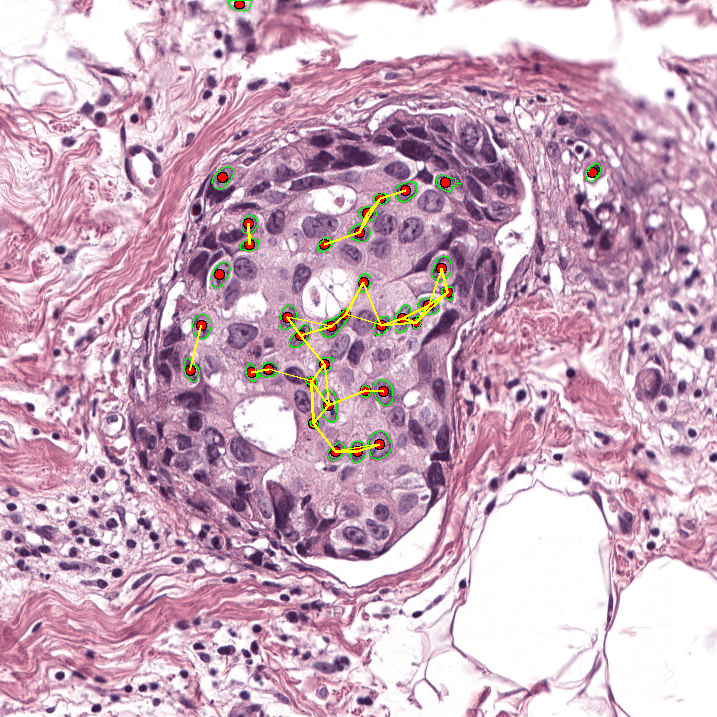}
    \caption{3-class explanation $\cg$}\label{fig:image3}
\end{subfigure}
\hfill
\begin{subfigure}{.24\linewidth}
    \centering
    \includegraphics[width=\linewidth]{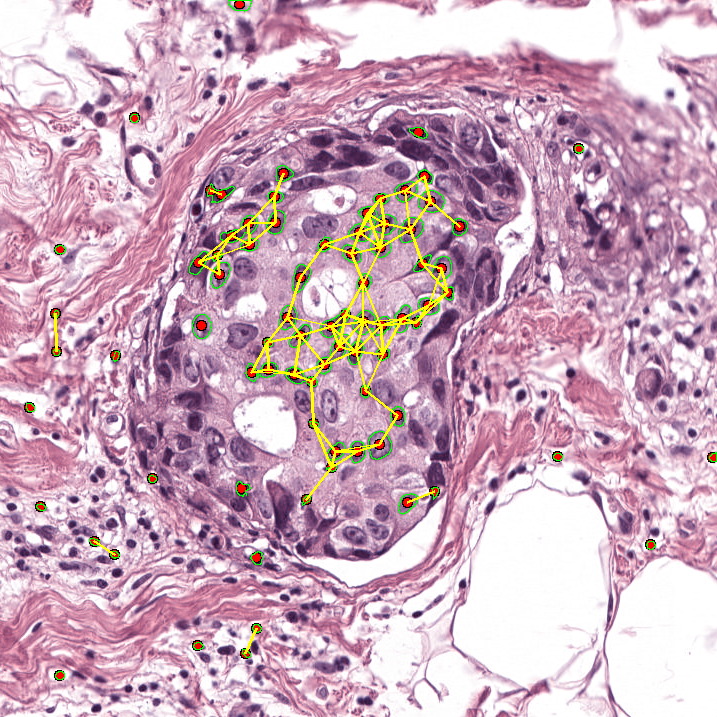}
    \caption{5-class explanation $\cg$}\label{fig:image4}
\end{subfigure}
{\caption{Qualitative comparison of original $\cg$ and $\cgexplainer$ $\cg$s for 2, 3 and 5-class scenarios for a DCIS $\troi$.}
\label{fig:troi_classes}}
\end{figure*}

\begin{figure}[t]
\centering
\setlength{\belowcaptionskip}{-10pt}
\includegraphics[width=\linewidth]{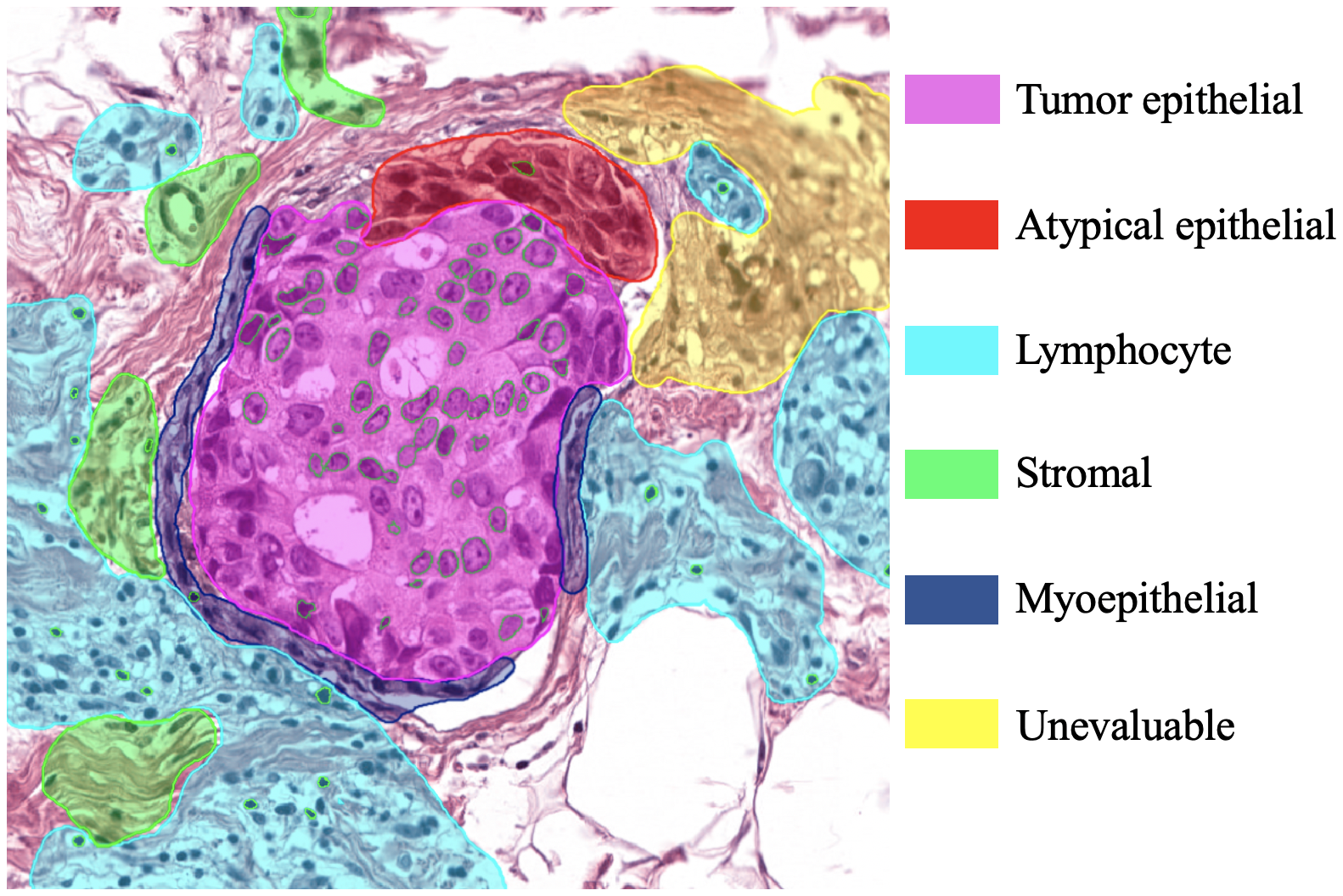}
\caption{Nuclei types annotation. Overlaid segmentation masks of nuclei from 5-class explanation in green.}
\label{fig:nuclei}
\end{figure}

We evaluate $\cgexplainer$ on BRACS dataset, an in-house collection of BReAst Carcinoma Subtyping\footnote{currently pending approval for releasing the dataset} images. 
The dataset consists of 2080 $\troi$s acquired from 106 H\&E stained breast carcinoma whole-slide-images (WSIs). The $\troi$s are extracted at $40\times$ magnification producing images of various sizes and appearances.
The $\troi$s are annotated by the consensus of three pathologists as: normal (N), benign\footnote{includes benign and usual ductal hyperplasia} (B), atypical\footnote{includes flat epithelial atypia and atypical ductal hyperplasia} (A), ductal carcinoma in situ (D), and invasive (I) (a 5-class problem). We also study two simplified scenarios: (1) a 2-class problem: benign (N+B) and malignant (D+I) categories, and (2) a 3-class problem: benign (N+B), atypical (A), and malignant (D+I) categories. These scenarios allow us to study the relation between the task complexity and the generated explanations. 
Non-overlapping train, validation and test splits are created at WSI-level consisting of 1356, 365, and 359 $\troi$s respectively.

\subsection{Implementation details}

The experiments are conducted using PyTorch~\cite{Paszke2019} and the DGL library~\cite{Wang2019}.
The $\cgnn$ consists of three GIN layers with a hidden dimension of $32$. Each GIN layer uses a 2-layer MLP with $\mathrm{ReLU}$ activation.
The classifier consists of a 2-layer MLP with $64$ hidden neurons that maps the hidden dimensions to the number of classes. The model is trained using the Adam optimizer with an initial learning rate of $10^{-3}$ and a weight decay of $5\times10^{-4}$. The batch size is set to 16. 

The explanation module uses the trained $\cgnn$. The mask $M_V$ is learned by using the Adam optimizer with a learning rate of $0.01$. 
The size constraint and the entropy constraint contribute to the loss by weighting factors $\alpha_{M_V} = 0.005$ and $\alpha_{\entropy} = 0.1$ respectively. 
The weights are adjusted such that the individual losses have comparable range.
An early stopping mechanism is triggered, if $G_s$ predicts a different label before reaching convergence. This ensures that the graph and its explanation  always have the same prediction.

\subsection{Quantitative and qualitative analyses}



We conduct absolute and comparative analyses between $\cgexplainer$ and random-explainer ($\rgexplainer$). $\rgexplainer$ generates a random explanation from an original $\cg$ for a $\troi$ by retaining equal number of nodes and edges as retained by $\cgexplainer$.
We quantitatively and qualitatively evaluate the explainers under 2, 3 and 5-class scenarios, and assess them using surrogate metrics in absence of ground truth explanations.
Table \ref{table:table1} presents the weighted F1-scores for $\cgnn$s, the average node and edge reduction in $\cgexplainer$ explanations, and cross-entropy ($\ce$) loss of $\cgnn$ for processing the original $\cg$, $ \cgexplainer$-based $\cg$ and $\rgexplainer$-based $\cg$. The cross-entropy is computed between the predicted logits and ground-truth labels of the $\troi$s.

The $\cgexplainer$ removes a large percentage of nodes and edges to generate compact explanations across 2, 3 and 5-class scenarios while preserving the $\troi$ predictions.
The decrease in the percentage of node reduction with the increase in the number of classes indicates that with the increment of task complexity, the explainer exploits more nodes to extract valuable information. A similar pattern is observed for the edge reduction.
Further, the reduction percentage decreases with the increase in the malignancy of the $\troi$. This indicates that the explainer discards abundantly available less relevant benign epithelial, stromal and lymphocytes, and retains relevant tumor and atypical nuclei.
Combining the $\cg$ explanations in Figure \ref{fig:troi_classes} and the nuclei types annotation in Figure \ref{fig:nuclei}, we infer that the explanations retain relevant tumor epithelial nuclei for DCIS diagnosis.
For 2-class scenario, the $\cg$ includes tumor nuclei in the central region of the gland. Few tumor nuclei are sufficient to differentiate (D) from (N+B).
For 3-class scenario, the $\cg$ includes more tumor nuclei in the central region and the periphery of the gland and does not consider atypical nuclei. This pattern differentiates (D) from (A).
For 5-class scenario, the $\cg$ includes more tumor nuclei distributed within and around the gland, and some lymphocytes around the gland. The $\cg$ also includes more cellular interactions to identify a large cluster of tumor nuclei. Pathologically this behavior differentiates (D) from (I) which has small clusters of tumor nuclei scattered throughout the $\troi$.
Additionally, the retained tumor nuclei and their interactions are consistent with increasing task complexity.


Further, we compare the class-wise logits for original, $\cgexplainer$ and $\rgexplainer$ $\cg$ via cross-entropy ($\ce$). Table \ref{table:table1} presents the class-wise $\ce$ and average $\ce$ across \emph{all} the classes. 
The $\cgexplainer$-based $\cg$ and the original $\cg$ have comparable class-wise $\ce$ and average $\ce$ across all scenarios. 
We observe that in each scenario, the $\rgexplainer$-based $\cg$ is biased towards one class. For instance, in the 2-class scenario, $\rgexplainer$ frequently predicts the class (N+B) leading to a per-class $\ce$ smaller than $\cgexplainer$. However, on average across \emph{all} the classes, the $\rgexplainer$ $\ce$ is consistently higher than the $\cgexplainer$. This conveys that the $\rgexplainer$ removes relevant entities from $\cg$s, thereby increasing the loss.
These qualitative and quantitative analyses conclude that the $\cgexplainer$ generates meaningful and consistent explanations.

\section{Conclusion}
We believe that our work, though preliminary, is a step in the right direction towards better representations and interpretability in DP. We have herein focused on the methodological introduction and cell-level analyses.
In future work, we plan to extend our approach to other biological entities and further to pathological assessment.
Ultimately, our goal is to understand any information additional to an ML model prediction that one needs to provide to a user, to build trust and to facilitate adoption and deployment of such ML technologies in clinical scanarios.

\section*{Acknowledgement}
We thank Nadia Brancati, Maria Frucci and Daniel Riccio, our  colleagues from Institute for High Performance Computing and Networking - CNR, Italy, for their insights and expertise that greatly assisted the research. We are also immensely grateful to 
Maurizio Do Bonito and Gerardo Botti from National Cancer Institute - IRCCS-Fondazione Pascale, Italy, and Giuseppe  De Pietro from Institute for High Performance Computing and Networking - CNR, Italy for their support and collaborative research with IBM Research Zurich, Switzerland.

{\small
\bibliographystyle{icml2020}
\bibliography{Histo-pathology.bib}
}

\end{document}